\documentclass{ifacconf}

\usepackage{graphicx}      
\usepackage{natbib}        

\usepackage{color}
\usepackage{url}
\usepackage{amsmath,amssymb,amsfonts,mathrsfs}
\usepackage{algorithmic}
\usepackage{booktabs}
\usepackage{multirow}
\usepackage{hhline}
\usepackage[utf8]{inputenc}
\usepackage[T1]{fontenc}
\usepackage{textcomp}
\usepackage{comment}
\usepackage{xcolor}
\usepackage{dsfont}
\usepackage{soul}
\usepackage{algorithm}
\usepackage{algorithmic}

\newtheorem{problem}{Problem}

\newtheorem{proof}{Proof}
\newtheorem{definition}{Definition}

\newcommand{\mR}{{\mathbb R}}

\newcommand{\mP}{{\mathbb P}}
\newcommand{\mU}{{\mathbb U}}

\newcommand{\bF}{{\mathbf F}}

\newcommand{\bQ}{{\mathbf Q}}
\newcommand{\bP}{{\mathbf P}}
\newcommand{\bR}{{\mathbf R}}

\newcommand{\bI}{{\mathbf I}}

\newcommand{\bX}{{\mathbf X}}

\newcommand{\bx}{{\mathbf x}}

\newcommand{\bg}{{\mathbf g}}

\newcommand{\bu}{{\mathbf u}}
\newcommand{\bU}{{\mathbf U}}

\newcommand{\bPhi}{{\mathbf \Phi}}

\newcommand{\bs}{{\mathbf s}}
\newcommand{\bff}{{\mathbf f}}
\newcommand{\bc}{{\mathbf c}}
\newcommand{\bC}{{\mathbf C}}

\newcommand{\bD}{{\mathbf D}}
\newcommand{\brho}{{\boldsymbol{\rho}}}

\begin{document}
\begin{frontmatter}

\title{Off-Road Navigation  of Legged Robots Using Linear Transfer Operators \thanksref{footnoteinfo}} 

\thanks[footnoteinfo]{This work was supported by the NSF under grant 2031573, 1932458, and 1942523.}

\author[asd]{Joseph Moyalan} 
\author[asd]{Andrew Zheng} 
\author[asd]{Sriram S.K.S Narayanan}
\author[asd]{Umesh Vaidya}

\address[asd]{Mechanical Engineering Department, Clemson University, Clemson, SC 29634 USA (e-mail:\{jmoyala,azheng,sriramk,uvaidya\}@clemson.edu).}

\begin{abstract}                
This paper presents the implementation of off-road navigation on legged robots using convex optimization through linear transfer operators. Given a traversability measure that captures the off-road environment, we lift the navigation problem into the density space using the Perron-Frobenius (P-F) operator. This allows the problem formulation to be represented as a convex optimization. Due to the operator acting on an infinite-dimensional density space, we use data collected from the terrain to get a finite-dimension approximation of the convex optimization. Results of the optimal trajectory for off-road navigation are compared with a standard iterative planner, where we show how our convex optimization generates a more traversable path for the legged robot compared to the suboptimal iterative planner. Simulation video of the implementation of our algorithm on legged robot is available at \url{https://youtu.be/T4sR22dVZa8}
\end{abstract}


\end{frontmatter}

\section{Introduction} \label{sec:introduction}
Legged robots have seen great success in traversing through off-road terrain \citep{miki2022learning}. In particular, the success can be attributed to the legged robot's versatile architectural and control framework for navigating through structured and unstructured environments \citep{tranzatto2022cerberus}. Although the capabilities of the legged system have been shown in off-road environments, the path traversed by the legged robot is not necessarily optimal. This is due to the off-road environment having an unclear distinction between traversable and non-traversable areas, giving rise to varying levels of traversability. Moreso, due to the wide range of traversability, the representation of the traversability is nonconvex, further increasing the difficulty of optimal navigation.

In most traditional navigation algorithms, the navigation problem does not deal with the unstructured nature of off-road terrains and instead solves the navigation problem in a well-structured environment \citep{lavalle2006planning}. These well-structured environments have a clear distinguishment between traversable and non-traversable regions. The success of these navigation problems in binary environments is seen through the plethora of algorithms that generate trajectories in obstacle-free regions, such as artificial potential methods, search algorithms, and sample-based methods \citep{khatib1986real,rimon1990exact,hart1968formal,lavalle1998rapidly,karaman2011sampling}. Artificial potential field-based methods represent an attractive well to pull trajectories into the target set and obstacles with a repulsive field to repel any trajectories heading towards the obstacle set. However, these methods do not extend to obstacles with varying degrees of traversability.

Algorithms that do generalize to navigation in the unstructured environment are typically through iterative local planning methods such as \citep{hart1968formal, lavalle1998rapidly}. These algorithms use cost maps to search for traversable regions, generating optimal feasible trajectories. For large-scale problems, these iterative planners add heuristic functions to guide the search so that the navigation problem is scalable. However, to guarantee optimality, the the heuristic function needs to be consistent and admissible, which are non-trivial to formulate in off-road environments. 

Recent works have shown that the navigation problem can alternatively be formulated in the dual space of density \citep{vaidya2018optimal}. The linear transfer operators such as Koopman and Perron-Frobenius (P-F) operators have been extensively studied in the literature to get a convex optimization framework for the optimal control problems \citep{moyalan2021sum,huang2022convex, moyalan2023data}. Similarly,  by utilizing Koopman and P-F operators, the navigation problem can be written as an infinite-dimensional convex optimization problem in the density space for structured environments \citep{yu2022data}. The infinite-dimensional optimization problem is then approximated to a finite-dimensional optimization problem using data \citep{huang2020data}. The use of a density-based navigation framework has also been explored for unstructured environments where the entire region can be riddled with obstacles with varying degrees of difficulty \citep{moyalan2023convex}. The main advantage of the convex framework for density-based navigation is obtaining a globally optimal solution while guaranteeing constraints.    

This work integrates the trajectory optimization method in \citep{moyalan2023convex} to find globally optimal traversable paths with a legged robot in simulation. We first show that the path generated is optimal with respect to traversability, and then showcase how well the legged robot traverses the path through control effort and state trajectory error. Furthermore, we compare with other planning algorithms and denote the sub-optimal traversable path by these planners. 

The rest of the paper is organized as follows. Section \ref{sec:preliminaries} discusses preliminaries to the operator theoretic approach for navigation. Section \ref{sec:robot_navigation} discusses planning stages of quadruped navigation. Then, Section Accordingly, Section \ref{sec:results} showcases the results of the navigation using the P-F operator on legged robots and conclusive remarks are made in Section \ref{sec:conclusion}.

\section{Preliminaries} \label{sec:preliminaries}
In this section, we provide several notations and definitions regarding the Koopman operator theory. We will also provide a method for data-driven approximation of Koopman operators.
\subsection{Notations and Definitions}
Consider the dynamics given as
\begin{equation}
    \dot{\bx} = \bF(\bx) \label{eq:dyna_sys}
\end{equation}
such that $\bx \in \bX \subset {\mR}^n$. We also assume that $\bF(\bx) \in \mathcal{C}^1(\bX,\mR^n)$, i.e. the space of continuously differentiable functions on $\bX$. Let us denote $\bs_t(\bx_0)$ to be the solution of a dynamical system (\ref{eq:dyna_sys}) at time $t$ starting from the initial point $\bx_0$. We denote $\mathcal{B}(\bX)$ to be Borel $\sigma$-algebra on $\bX$ and $\mathcal{M}(\bX)$ to be the vector space of real-valued measure on ${\cal B}(\bX)$. We also denote $\mathcal{L}_{\infty}(\bX)$ and $\mathcal{L}_1(\bX)$ to be the space of essentially bounded and integrable functions on $\bX$ respectively. Next, we will define the Koopman operator and the Perron-Frobenius (P-F) operator \citep{lasota1998chaos}.
\begin{definition}[Koopman Operator]$\mathscr{U}_t:{\cal L}_{\infty}(\bX)\;\rightarrow\;{\cal L}_{\infty}(\bX)$ for (\ref{eq:dyna_sys}) is defined as
\begin{align}
    [\mathscr{U}_t \varphi](\bx_0) = \varphi(\bs_t(\bx_0)), \label{eq:def_Koopman_operator}
\end{align}
where $\varphi \in \mathcal{L}_{\infty}(\bX) \bigcap \mathcal{C}^1(\bX)$ is a scalar function. The infinitesimal generator for the Koopman operator is given by
\begin{align}
    \lim_{t\to 0}\frac{[\mathscr{U}_t\varphi](\bx_0)-\varphi(\bx_0)}{t}={\bF}(\bx_0)\cdot \nabla \varphi(\bx_0)=:\mU_{\bF} \varphi .\label{eq:def_Koopman_generator}
\end{align}
\end{definition}
\begin{definition}[Perron-Frobenius Operator] $\mathscr{P}_t:{\cal L}_1(\bX)\to {\cal L}_1(\bX)$ for (\ref{eq:dyna_sys}) is defined as
\begin{align}
    [\mathscr{P}_t\psi](\bx_0) = \psi(\bs_{-t}(\bx_0)) \left| \frac{\partial \bs_{-t}(\bx_0)}{\partial \bx}\right|, \label{eq:def_PF_operator}
\end{align}
where $|.|$ stands for the determinant and $\bs_{-t}(\bx_0)$ represents the solution of a dynamical system (\ref{eq:dyna_sys}) at time $t$ moving backwards starting from the initial point $\bx_0$. The infinitesimal generator for the P-F operator is given by
\begin{align}
    \lim_{t\to 0}\frac{[\mathscr{P}_t\psi](\bx_0)-\psi(\bx_0)}{t}= -\nabla \cdot (\bF(\bx_0)\psi(\bx_0))=:\mP_{\bF} \psi .\label{eq:def_PF_generator}
\end{align}
\end{definition}
We use Koopman and P-F operators to lift the finite-dimensional nonlinear dynamics from state space to infinite-dimensional space of functions. We utilize the positivity and Markov property of the two operators for the finite-dimension approximation of these operators.
\begin{definition}\label{def_aeuniformstable}[Almost everywhere (a.e.) stability]  The equilibrium point or an attractor set of the system (\ref{eq:dyna_sys}) represented by $A$ is said to be a.e. stable w.r.t. measure $\mu_0\in {\cal M}(\bX)$ if 
\begin{eqnarray*}
\mu_0\{\bx\in \bX: \lim_{t\to \infty} \bs_t(\bx)\notin A\}=0.\label{eq_aeunifrom}
\end{eqnarray*}
\end{definition}
\subsection{Data-driven approximation of Koopman and P-F operators}\label{ASDMD}
Several methods exist in the literature for the approximation of Koopman and P-F  operator matrices \citep{schmid2010dynamic,williams2015data,huang2018data}. In this paper, we make use of the Approximate naturally structured dynamic mode decomposition (ANSDMD) \citep{moyalan2023convex}, which helps us to gain numerical efficiency by preserving the positivity and Markov properties of these operators approximately. In this method, the snapshots of time-series data of a continuous-time dynamical system are arranged as follows:
\begin{align*}
    \mathscr{X}= [\bx_1,\bx_2,\ldots,\bx_P],\;\;\;\;\mathscr{Y} = [\mathbf{y}_1,\mathbf{y}_2,\ldots,\mathbf{y}_P] 
\end{align*}
where $\bx_i\in \bX$ and $\mathbf{y}_i\in \bX$ such that $\mathbf{y}_i=\bs_{\Delta t}(\bx_i)$. Let ${\bPhi}=[\phi_1,\ldots,\phi_Q]^\top$ be the set of basis functions. Now, the finite-dimensional approximation of the Koopman operator ($\mathcal{U}$) can be obtained by solving the following least square problem. 
\begin{equation*}\label{edmd_op}
\min\limits_{\mathcal{U}}\parallel {\bf G}\mathcal{U}-{\bf A}\parallel_F,
\end{equation*}
\begin{eqnarray*}\label{edmd1}
{\bf G}=\frac{1}{P} \bPhi( \mathscr{X}) \bPhi( \mathscr{X})^\top,\;\;\;
{\bf A}=\frac{1}{P} \bPhi( \mathscr{X}) \bPhi( \mathscr{Y})^\top,
\end{eqnarray*}
with $\mathcal{U},{\bf G},{\bf A}\in\mathbb{R}^{Q\times Q}$, $\parallel \cdot \parallel_F$ stands for Frobenius norm. We assume that the basis functions selected are positive. Next, we perform row normalization on $\mathcal{U}$ to get $\hat{\mathcal{U}}$ which helps us to avoid Markov constraints.
 We obtain the P-F matrix as  
$\hat{\mathcal{P}} = \hat{\mathcal{U}}^{\top}$. The generator of the P-F operator is obtained as
\begin{eqnarray*}
{\mP}_{\bff}\approx \frac{\hat{\mathcal{P}}-{\bf I}}{\Delta t}=:\bP.\label{PF_approximation}
\end{eqnarray*}
where $\bI$ is the identity matrix.
\begin{figure*}[t!]
    \centering
    \includegraphics[width=\linewidth]{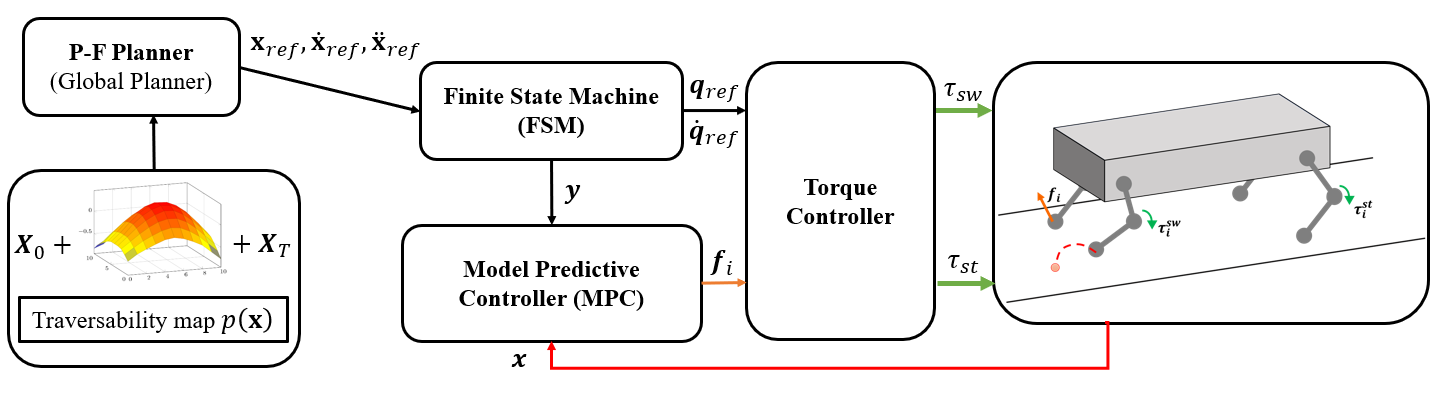}
    \caption{The different planning stages for the quadrupedal robot.}
    \label{fig:flow_chart}
\end{figure*}
\section{Planning for Quadruped Navigation}\label{sec:robot_navigation}
This section discusses the various stages of planning for quadruped navigation as shown in Figure \ref{fig:flow_chart}. The navigation problem is first defined by a global planner which provides a reference trajectory ($\bx_{ref}$) connecting the initial set $\bX_0$ and the target set $\bX_T$. Sequentially, a finite state machine (FSM) handles the contact scheduling, footstep locations while the model predictive controller (MPC) optimizes the forces to track the reference trajectory. Finally, a leg controller converts the MPC and the finite state machine output into joint space torque control. We note that this paper's main contribution is the usage of the P-F planner for global trajectory planning for off-road terrain. The P-F planner incorporates Koopman operator theory to get a convex optimization, where the following solution is a feedback controller used as an optimal reference trajectory ($\bx_{ref}$) for traversing the unstructured environment.

\subsection{P-F planner}
In this section, we describe the design of a convex optimization framework for off-road navigation using the traversability information. The traversability description of the terrain describes terrain information such as the elevation map, slope, terrain texture, and terrain roughness. We define a nonnegative scalar function $p(\bx): \mathbb{R}^n \to [0,1]$ in the space of integrable functions to represent the traversability map. The traversability map design involves using onboard sensors on the quadruped robot such as cameras, LIDAR, and IMU \citep{huang2021decentralized,armbrust2011ravon,bagheri2017development}. In this paper, we design a feedback control input $\bu(\bx)\in \bU\subset \mR^m$ to navigate the quadrupedal robot from some initial state $\bX_0  \subset \bX$ to some final target set $\bX_T  \subset \bX$ while minimizing the traversability function.
Therefore, we consider the following cost function.
 \begin{equation}
J(\mu_0)=\min_{\bu}\;\int_{\bX}\int_0^\infty  p(\bx(t))+ p(\bx(t))\|\bu\|_1dt\;d\mu_0 \label{eq:control_cost}
\end{equation}
where $\mu_0$ is the measure capturing the distribution of the initial state. The second term in \eqref{eq:control_cost} minimizes cost with respect to control input. In \eqref{eq:control_cost}, we minimize the cost function averaged over all initial states $\bx\in \bX_0$ instead of minimizing over $\bx\in \bX$. This is done by weighting the cost function with $\mu_0$. Hence, different initial conditions are weighted differently. In this paper, we use the dynamics of a single integrator in the navigation problem since we are only interested in getting a reference trajectory, but in general, any nonlinear dynamics can be incorporated into the navigation problem. The optimal off-road navigation problem is given below.

\begin{problem}\label{problem_statement2} Find a feedback control $\bu(\bx)$ to navigate almost every system trajectory for  $\dot{\bx}=\bu(\bx)$ starting from the initial set $\bX_0$ to the target set $\bX_T$ such that the following cost is minimized. 
\begin{subequations}\label{problem_B}
    \begin{alignat}{5}
    \min_{\bu}&\;\;\int_{\bX} \left(\int_0^{\infty}p(\bx(t))+ p(\bx(t))\|\bu(t)\|_1\;dt\right) d\mu_0\label{problem_a}\\
    {\rm s.t.}&\;\;\; \dot \bx =\bu,\;\;\;\;\lim_{t\to \infty} \bx(t)\in \bX_T. \label{problem_b}
       \end{alignat}
\end{subequations}
 The solution to the off-road navigation problem is the most traversable path to the target set. 
\end{problem}
The convex optimization framework for Problem \ref{problem_statement2} in terms of optimization variables 
$\rho \in{\cal L}_1(\bX_1)\cap {\cal C}^1(\bX_1,\mR_{\geq 0})$ and $\bar \brho\in{\cal C}^1(\bX_1,\mR)$ for the optimal off-road navigation problem can be written as follows \citep{moyalan2022data1}: 
\begin{subequations}\label{convex_formulation}
\begin{align}
    &\inf_{\rho,\bar \brho}\int_{\bX_1} p(\bx)\rho(\bx)+p(\bx) \; \|\bar{\brho}(\bx)\|_1 \;d\bx \label{ocp_dens}\\
&{\rm s.t}.\;\;\;
\nabla\cdot\bar{\brho}=h_0,\label{feas221}
\end{align}
\end{subequations}
where $h_0(\bx)=\frac{d \mu_0}{d \bx}$. The optimal feedback control input is obtained from the solution to the above optimization problem as follows:
\begin{align}
    \bu^\star(\bx) = \frac{\bar{\brho}^\star(\bx)}{\rho^\star(\bx)} \label{optimal_control}
\end{align}
where $(\rho^\star,\bar{\brho}^\star)$ are the solution of \eqref{convex_formulation}.

The finite-dimension approximation of \eqref{convex_formulation} can be done by selecting a finite number of basis functions. The finite-dimension approximation of \eqref{feas221} is done by applying the definition of P-F generator \eqref{eq:def_PF_generator} in ANSDMD algorithm as described in Section \ref{ASDMD}.  We use radial basis functions for our basis. Let $\Phi(\bx) = [\phi_1(\bx),\phi_2(\bx),\dots,\phi_Q(\bx)]^\top$ be the set of basis functions. Next, we define $\bar{\brho}(\bx) = [\bar{\rho}_1(\bx),\dots,\bar{\rho}_m(\bx)]^\top$ and  $\mathbf{\Gamma}(\bx) = [\Gamma_1(\bx),\dots,\Gamma_m(\bx)]^\top $. Now, the following terms can be expressed as combinations of basis functions.
\begin{align}
    h_0 = \bPhi^\top {\bf t},\; \rho = \bPhi^\top {\bf r},
    \;{\bar{\rho}}_j = \bPhi^\top {\bf z}_j\;\Gamma_j(\bx) = \bPhi^\top \textbf{w}_j. \label{term_approx}
\end{align}
where $|\bar{\rho}_j(\bx)|< \Gamma_j(\bx)$. Minimizing $\Gamma_j(\bx)$ will indirectly minimize $|\bar{\rho}_j(\bx)|$. Replacing  $\|\bar{\brho}(\bx)\|_1$ by $\sum_j\Gamma_j(\bx)$ in \eqref{ocp_dens}, we perform the following finite-dimensional approximation.
\begin{subequations}
\begin{align}
    \inf_{\rho,\bar \brho}&\int_{\bX_1}\left[p(\bx) \rho(\bx)+p(\bx) \; \sum_j\Gamma_j(\bx) \;\right]d\bx \approx \nonumber\\
    &\underset{\textbf{w}_j,{\bf z}_j,{\bf r}}{\min}\;{\bf B_1}^\top {\bf r}+{\bf B_2}^\top \sum_j\textbf{w}_j \nonumber\\
    & \text{s.t.}~~   \textbf{w}_j - {\bf z}_j > \mathbf{0} \nonumber\\
    & \quad\quad \textbf{w}_j + {\bf z}_j > \mathbf{0} 
\end{align}
\end{subequations}
where ${\bf B}_1 = \int_{\bX_1}p(\bx) \; \Phi(\bx) \;d\bx$ , ${\bf B}_2 = \int_{\bX_1}p(\bx) \; \Phi(\bx) \;d\bx$ and $\mathbf{0}$ is a zero vector.
Henceforth, the finite-dimensional approximation of \eqref{convex_formulation} is written as 
\begin{eqnarray}
&\underset{\textbf{w}_j,{\bf z}_j,{\bf r}}{\min}\;{\bf B_1}^\top {\bf r}+{\bf B_2}^\top \sum_j\textbf{w}_j \nonumber\\
\text{s.t.} &\textbf{w}_j - {\bf z}_j > \mathbf{0} \nonumber\\
&\textbf{w}_j + {\bf z}_j > \mathbf{0} \nonumber\\
&-\bP_{\bff}{\bf r}-\sum_j \bP_{\bg_j}{\bf z}_j=\bf t\nonumber
\end{eqnarray}
\subsection{Model Predictive Control (MPC) for Quadruped}
The control input for the quadruped robot is obtained using a nonlinear MPC. The MPC uses a quadratic objective function to track the reference states $\bx_{k+1, ref}$ and control reference trajectory $\bu_{q,k, ref}$ obtained from the P-F planner. The MPC objective function is given as follows:

\begin{align}
    \min_{\bx, \bu} \hspace{2mm} &\sum_{k=0}^{N-1} ||\bx_{k+1} - \bx_{k+1, ref} ||_{\bQ_k} + ||\bu_{q,k} - \bu_{q,k,ref} ||_{\bR_k} \\
    \textrm{subject to} \nonumber \\
    & f(\bx_{k}, \bx_{k+1}, \bu_{q,k}) = 0 \label{eq:dynamics} \\
    & \underline{\bc}_k \leq \bC_k \bu_{q,k} \leq \bar{\bc}_k \label{eq:constraint1} \\
    & \bD_k\bu_{q,k} = \mathbf{0}
\end{align}

where $\bx_k$ and $\bu_{q,k}$ are the state and control decision variables entering the MPC, $f$ is the dynamic constraints, $N$ represents the receding horizon length, $\bQ_k \ge0$ represents state cost matrix, $\bR_k>0$ represents the control weight matrix, $\bC_k$ are the constraint matrix for each leg in contact, $\underline{\bc}_k$ and $\bar{\bc}_k$ are the upper and lower bounds on friction, and $\bD_k$ is the contact selection matrix.


\subsection{Torque Controller}
Based on the desired gait cycle, the finite state machine outputs a switching condition to determine whether a given foot is in a swing or stance phase. For the swing phase, an inverse dynamics-based controller is used to obtain the corresponding joint torques $\mathbf{\tau}^{sw}_i \in \mathbb{R}^l$ where $l$ is the number of joints in each leg. Using the reference trajectory for joint positions $\mathbf{q}_{i,ref} \in \mathbb{R}^l$ and velocities $\dot{\mathbf{q}}_{i,ref} \in \mathbb{R}^l$, the torque for each joint of the swing leg $i$ can be computed using inverse kinematics.
\begin{align} \label{eq:swing}
    &\mathbf{\tau}_i^{sw} = M_i(\mathbf{q}_i)\ddot{\mathbf{q}}_i + h_i(\mathbf{q}_i,\dot{\mathbf{q}}_i) \nonumber \\
    &+ K_p^{sw}(\mathbf{q}_{i, ref}-\mathbf{q}_i) + K_d^{sw}(\dot{\mathbf{q}}_{i, ref}-\dot{\mathbf{q}}_i)
\end{align}
where, for each leg $i$, $ (\mathbf{q}_i,\dot{\mathbf{q}}_i,\ddot{\mathbf{q}}_i) \in \mathbb{R}^l$ represent the joint position, velocity and acceleration respectively, $M_i \in \mathbb{R}^{l \times l}$ is the inertia matrix and $h_i: \mathbb{R}^l \xrightarrow{} \mathbb{R}^l$ represents the Coriolis and gravity effects acting on the leg and $(K_p^{sw}, K_d^{sw}) \in \mathbb{R}^{l \times l}$ are constant gain matrices.  
For the stance phase, the joint torques $\mathbf{\tau}^{st}_i$ is obtained by converting the foot forces $\bff_i$ from the MPC to joint space as follows
\begin{equation}
    \mathbf{\tau}_i^{st} = J_i^\top R_i^\top \mathbf{f}_i \label{tau_stance}
\end{equation}
where, $J_i \in \mathbb{R}^{l \times l}$ is the leg Jacobian for the $i$th feet in stance and $R_i \in {l \times l}$ is the corresponding rotation matrix.

\section{Results} \label{sec:results}
In this section, we demonstrate how we incorporate our off-road navigation using P-F operator. First, we discuss the simulation setup to incorporate the navigation problem for legged robots. Then, the following results are demonstrated on a quadruped in simulation using a physics simulator, Gazebo.
\begin{figure}[bp]
  \centering
  \includegraphics[width=1\linewidth]{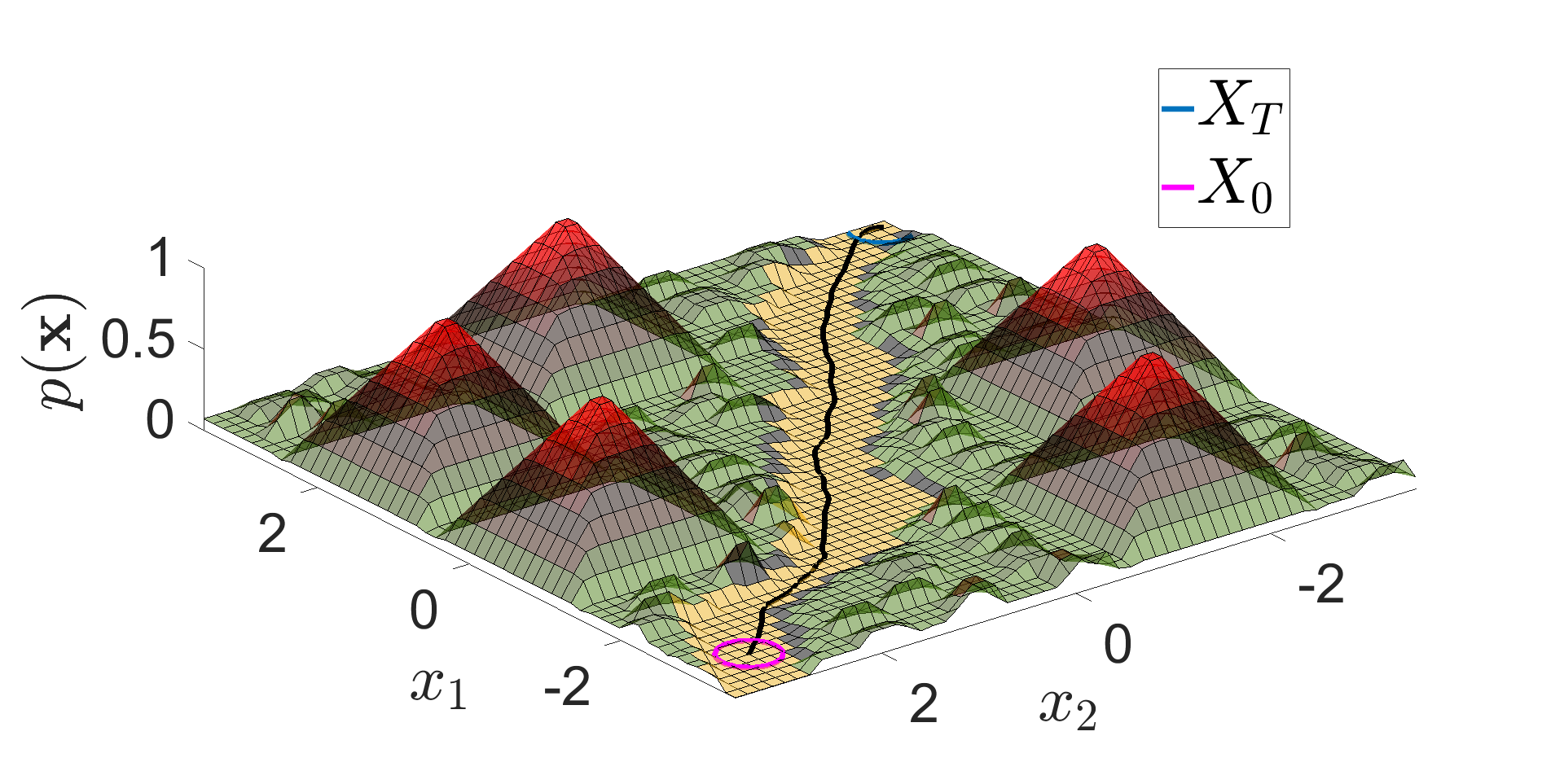}
\caption{Mountain pass environment in which P-F planner gives trajectory from $\bX_0$ to $\bX_T$ of road. }
\label{fig:terrain_plot}
\end{figure}
\subsection{Simulation Setup}
To incorporate the off-road navigation problem, we use the MATLAB CVX toolbox with Gurobi for our convex problem formulation. Furthermore, to include the trajectory from the optimization into a legged system, we devised a legged planning framework using terrain information to map the states from a reference trajectory of a particle to a reference trajectory of a rigid body with terrain and contact information. The performance of the off-road navigation is validated with the help of an open-source legged robot framework \citep{norby2022quad}.
\begin{figure*}[t!]
    \centering
    \includegraphics[width=\linewidth]{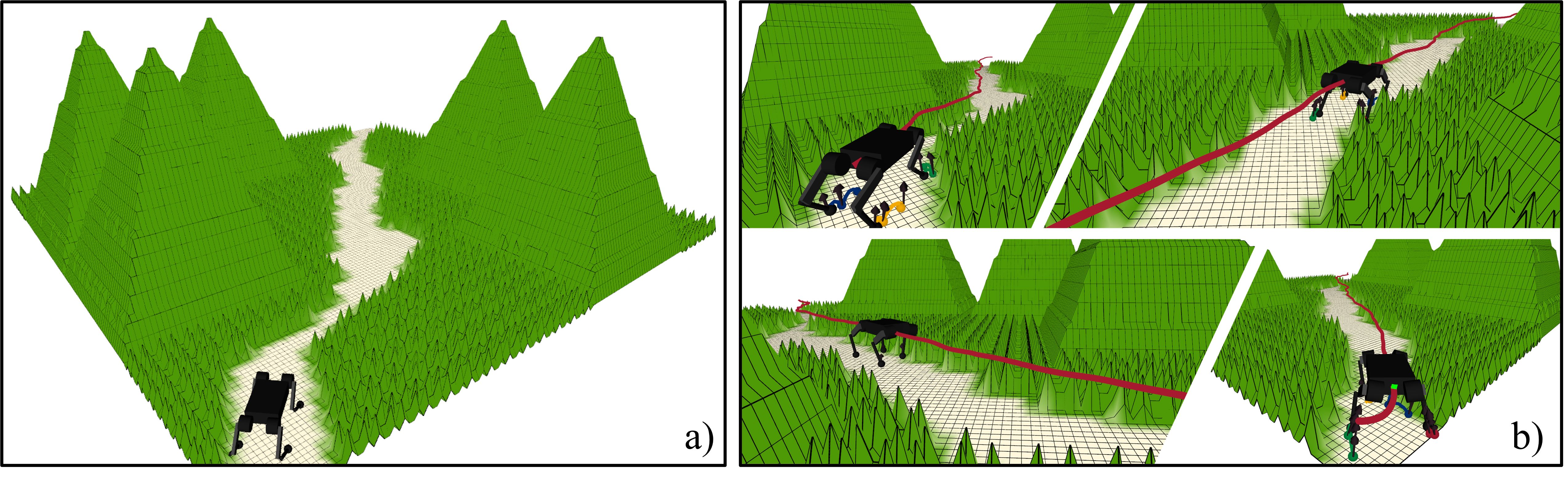}
    \caption{Quadruped navigating a mountain-pass. The optimal P-F planner generates a trajectory through the road, the region that is the most traversable. a) Isometric view of the terrain b) Legged robot navigating from $\bX_0$ to $\bX_T$}
    \label{fig:quad_nav_mountain_pass}
\end{figure*}

\subsection{Legged Navigation}
For the following off-road navigation problem shown in Figure \ref{fig:terrain_plot}, we define the traversability function $p(\bx) = \frac{h(\bx)}{h_{max}}$, where $h(\bx)$ is the terrain height and $h_{max}$ is the maximum terrain height, defining the traversability measure for our navigation problem in \eqref{problem_B}. Note, $p(\bx)$ is assumed to be nonnegative. From the following environment, Figure \ref{fig:quad_nav_mountain_pass} shows the optimized trajectory tracked on a quadruped. The path taken starts at the beginning of the mountain pass, $\bX_0$, and traverses on the road toward the end of the mountain pass, $\bX_T$. It is trivial to see here that the optimal solution to the off-road navigation problem is the trajectory $\bx(t)$ such that $\int_0^{\infty}p(\bx)\rho(\bx) + p(\bx)\sum_j\Gamma_j(\bx)dt = 0$, validating the optimality of the operator theoretic approach.

\begin{figure}[htbp]
  \centering
  \includegraphics[width=0.99\linewidth]{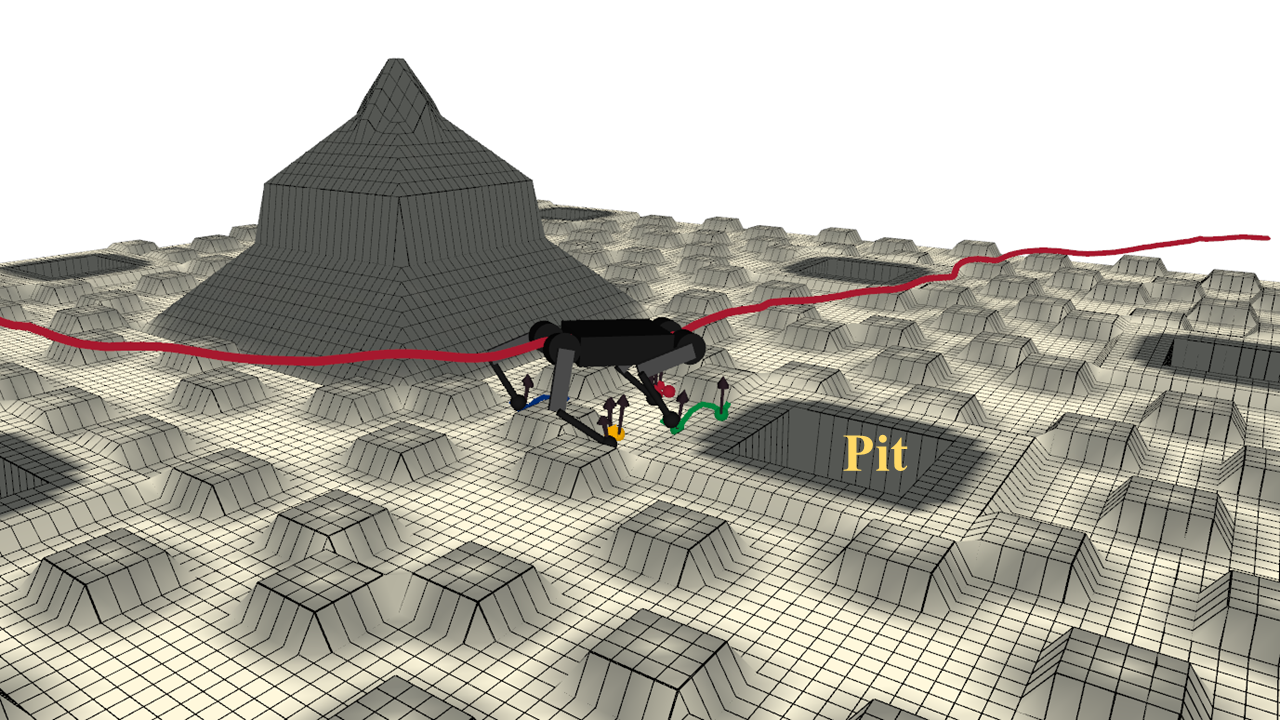}
\caption{Quadruped traversing through an environment with hard obstacles (pit) and unstructured terrains.}
\label{fig:mountain_pit}
\end{figure}
\begin{figure*}[t!]
  \centering
  \includegraphics[width=1\linewidth]{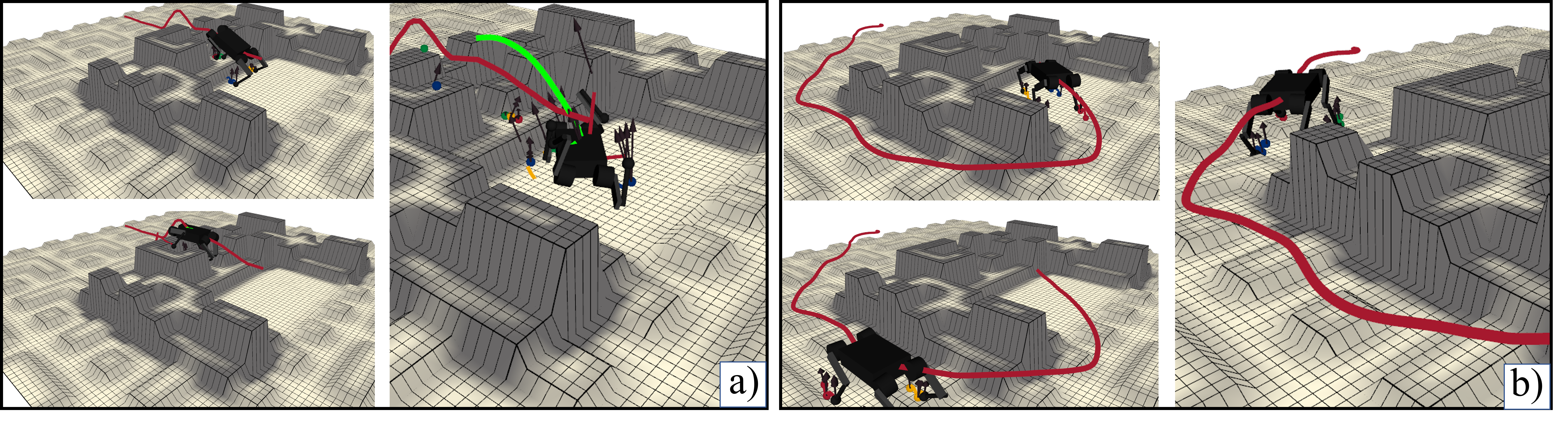}
\caption{Legged Navigation on Non-trivial unstructured terrain with a) $A^{\star}$ algorithm and b) P-F planner. The traversability cost along the trajectory for $A^{\star}$ and P-F planner are 1.03 and 0.75 respectively.}
\label{fig:c_shape_comparison}
\end{figure*}
\subsection{Legged Navigation on Terrain with Obstacles}
Correspondingly, we look at the navigation problem with unstructured terrain and obstacles as shown in Figure \ref{fig:mountain_pit}. For the following environment, we classify the hard obstacles using the following constraint condition
\begin{align}
    \int_{\bX_1}\mathds{1}_{\bX_b}(\bx)\rho(\bx)d\bx=0 \label{hard_constraint}
\end{align}
where $\bX_b \subset \bX_1$ represent the obstacle set and $\mathds{1}_{\bX_b}(\bx)$ represents the indicator function with respect to $\bX_b$. By adding  \eqref{hard_constraint} in the convex optimization problem given in \eqref{convex_formulation}, we obtain a feedback control such that the corresponding trajectory avoids the obstacle set while navigating optimally through the unstructured terrain. The density function $\rho(\bx)$ obtained as the solution to this modified optimization problem will have almost zero value inside the obstacle set. Figure \ref{fig:mountain_pit} showcases the quadruped navigating an environment with unstructured terrain and obstacles where obstacles are present in the form of pits. 

\subsection{Legged Navigation on Non-Trivial Terrains}
Lastly, we look at a non-trivial off-road experiment shown in Figure \ref{fig:c_shape_comparison}. The following environment consists of an unstructured terrain environment with a C-shaped terrain. To validate the performance of our off-road navigation algorithm using P-F planner, we compare the results to the A$^\star$ algorithm.

Note, designing a heuristic function for A$^\star$ that is admissible and consistent is difficult for off-road environments. As such, a Euclidean distance is selected as the heuristic function. Figure \ref{fig:c_shape_comparison} shows that the iterative path planner searches suboptimally across the terrain environment. As a result, the quadruped tries to track the reference trajectory but fails to navigate across the C-shaped terrain.

Accordingly, Figure \ref{fig:c_shape_comparison}b shows the navigation problem for our P-F planner. Since our convex optimization searches for a globally optimal solution, the generated trajectory goes around the C-shaped terrain. As shown, the globally optimal solution from the initial to the target set is traversable by the quadruped.


\section{Conclusion} \label{sec:conclusion}
In this work, we show the integration of the optimal navigation problem from a linear transfer operator approach on a legged system. First, we demonstrate the navigation problem as a convex optimization in the dual space of density. We then validate these results on a quadruped in a mountain pass environment, non-trivial unstructured terrains, and terrains with obstacles. Furthermore, we compare our optimization to a widely used search algorithm, A$^{\star}$, and demonstrate how our convex optimization is able to generate optimal traversable paths for the legged robot in comparison with the suboptimal trajectory generated from A$^{\star}$. Future works will modify our operator theoretic convex optimization for real-time performances and implement our existing algorithms onto a legged robot on hardware.


\bibliography{ref}             

\end{document}